\documentclass[sigconf]{acmart}
\AtBeginDocument{%
  }

\setcopyright{none}
\acmConference[KDD Workshop '25]{Workshop on Structured Knowledge for Large Language Models (SKnowLLM) at KDD 2025}{August 03--07,2025}{Toronto, Canada}
\usepackage{times}
\usepackage{latexsym}

\usepackage[T1]{fontenc}
\usepackage[utf8]{inputenc}

\usepackage{microtype}

\usepackage{graphicx}

\usepackage{enumitem}
\usepackage{listings}
\usepackage[most]{tcolorbox}

\usepackage{colortbl} % Add to your preamble if not already present
\definecolor{training_row}{RGB}{185, 235, 255}
\definecolor{eval_row}{RGB}{255, 219, 187}

\definecolor{structure_row}{RGB}{185, 235, 255}
\definecolor{input_row}{RGB}{255, 219, 187}
\definecolor{enterprise_row}{RGB}{190, 255, 190}
\definecolor{stringcolor}{rgb}{0.75,0.2,0.2}  % String color
\definecolor{keycolor}{rgb}{0.1,0.1,0.6}      % Key color
\definecolor{backgroundyellow}{RGB}{255, 245, 220}
\definecolor{mediumgray}{gray}{0.6} % A shade between gray (0.5) and lightgray (0.75)

\definecolor{Red}{RGB}{220, 38, 38}    % A bright red
\definecolor{Green}{RGB}{34,139,34}

\lstdefinelanguage{json}{
    basicstyle=\ttfamily\tiny,
    numbers=left,
    numberstyle=\tiny\color{gray},
    stepnumber=1,
    numbersep=3pt,
    showstringspaces=false,
    breaklines=true,
    frame=none,
    backgroundcolor=\color{backgroundyellow},
    keywordstyle=\color{keycolor},
    stringstyle=\color{stringcolor},
    morestring=[s]{"}{"},
    morecomment=[l]{//},
    commentstyle=\color{gray}
}

\newtcbox{\jsonbox}{on line, colback=background, colframe=gray!50, boxrule=0.5pt, arc=4pt, boxsep=2pt, left=2pt, right=2pt, top=2pt, bottom=2pt}

\begin{document}

%%
%% The "title" command has an optional parameter,
%% allowing the author to define a "short title" to be used in page headers.

\title{Fine-Tune an SLM or Prompt an LLM? The Case of Generating Low-Code Workflows}
%%
%% The "author" command and its associated commands are used to define
%% the authors and their affiliations.
%% Of note is the shared affiliation of the first two authors, and the
%% "authornote" and "authornotemark" commands
%% used to denote shared contribution to the research.
% \author{Ben Trovato}
% \authornote{Both authors contributed equally to this research.}
% \email{trovato@corporation.com}
% \orcid{1234-5678-9012}
% \author{G.K.M. Tobin}
% \authornotemark[1]
% \email{webmaster@marysville-ohio.com}
% \affiliation{%
%   \institution{Institute for Clarity in Documentation}
%   \city{Dublin}
%   \state{Ohio}
%   \country{USA}
% }

\author{Orlando Marquez Ayala}
\affiliation{%
    \institution{ServiceNow}
}
\email{orlando.marquez@servicenow.com}

\author{Patrice Bechard}
\affiliation{%
    \institution{ServiceNow}
}
\email{patrice.bechard@servicenow.com}

\author{Emily Chen}
\affiliation{%
    \institution{ServiceNow}
}
\email{emily.chen@servicenow.com}

\author{Maggie Baird}
\affiliation{%
    \institution{ServiceNow}
}
\email{maggie.baird@servicenow.com}

\author{Jingfei Chen}
\affiliation{%
    \institution{ServiceNow}
}
\email{jingfei.chen@servicenow.com}

% \author{Charles Palmer}
% \affiliation{%
%   \institution{Palmer Research Laboratories}
%   \city{San Antonio}
%   \state{Texas}
%   \country{USA}}
% \email{cpalmer@prl.com}

% \author{John Smith}
% \affiliation{%
%   \institution{The Th{\o}rv{\"a}ld Group}
%   \city{Hekla}
%   \country{Iceland}}
% \email{jsmith@affiliation.org}

% \author{Julius P. Kumquat}
% \affiliation{%
%   \institution{The Kumquat Consortium}
%   \city{New York}
%   \country{USA}}
% \email{jpkumquat@consortium.net}

%%
%% By default, the full list of authors will be used in the page
%% headers. Often, this list is too long, and will overlap
%% other information printed in the page headers. This command allows
%% the author to define a more concise list
%% of authors' names for this purpose.
% \renewcommand{\shortauthors}{Trovato et al.}

%%
%% The abstract is a short summary of the work to be presented in the
%% article.
\begin{abstract}
Large Language Models (LLMs) such as \mbox{GPT-4o} can handle a wide range of complex tasks with the right prompt. As per token costs are reduced, the advantages of fine-tuning Small Language Models (SLMs) for real-world applications --- faster inference, lower costs --- may no longer be clear. In this work, we present evidence that, for domain-specific tasks that require structured outputs, SLMs still have a \textit{quality} advantage. We compare fine-tuning an SLM against prompting LLMs on the task of generating low-code workflows in JSON form. We observe that while a good prompt can yield reasonable results, fine-tuning improves quality by 10\% on average. We also perform systematic error analysis to reveal model limitations.
\end{abstract}

%%
%% The code below is generated by the tool at http://dl.acm.org/ccs.cfm.
%% Please copy and paste the code instead of the example below.
%%


%%
%% Keywords. The author(s) should pick words that accurately describe
%% the work being presented. Separate the keywords with commas.
%% A "teaser" image appears between the author and affiliation
%% information and the body of the document, and typically spans the
%% page.
% \begin{teaserfigure}
%   \includegraphics[width=\textwidth]{sampleteaser}
%   \caption{Seattle Mariners at Spring Training, 2010.}
%   \Description{Enjoying the baseball game from the third-base
%   seats. Ichiro Suzuki preparing to bat.}
%   \label{fig:teaser}
% \end{teaserfigure}

\received{30 May 2025}
% \received[revised]{12 March 2009}
% \received[accepted]{5 June 2009}

\settopmatter{printacmref=false} % Removes ACM reference format

%%
%% This command processes the author and affiliation and title
%% information and builds the first part of the formatted document.
\maketitle

\section{Introduction}
Generative AI is pushing the boundaries of AI-based software. As Large Language Models (LLMs) become more capable, they have become an integral part of the software stack. Although prompting state-of-the-art models such as \mbox{GPT-4o} \cite{hurst2024gpt}, Gemini 2.0 Flash \cite{team2023gemini}, or Claude 3.5 Sonnet \cite{Claude3S} generally suffice for traditional tasks like question-answering or summarization \cite{fu-etal-2024-tiny}, it is not always the case on domain-specific tasks \cite{make6010018}.

While building \textit{Flow Generation}, an application that generates low-code workflows based on textual user requirements, we explored whether fine-tuning a language model was necessary to achieve the desired software quality. To reduce infrastructure needs, we only considered Small Language Models (SLMs), having less than 15 billion parameters.

\begin{figure}[t]
    \centering
    \includegraphics[width=0.9\linewidth]{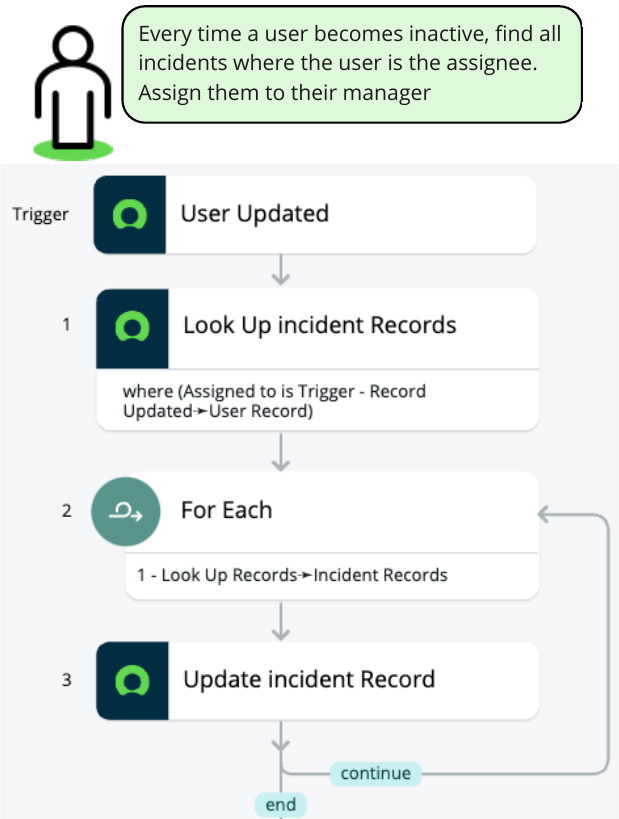}
    \caption{Task consists of generating a complete workflow from a user requirement.}
    \Description{Task consists of generating a complete workflow from a user requirement.}
    \label{fig:sample_flow}
    \vspace{-0.3cm}
\end{figure}

Enterprise workflows are a common way to automate repetitive yet complex processes. They are represented as a series of steps that are executed under certain conditions to fulfill specific goals. While low-code workflows tend to be built in easy-to-use interfaces, creating them still requires expert knowledge of the enterprise system. Figure \ref{fig:sample_flow} shows an example of a simple four-step workflow. Some of the challenges in this task are:
\begin{itemize}
    \item Every step must exist in the system environment; they vary by installation and users can add custom steps.
    \item Workflow structure must follow a set of well-defined rules, using concepts such as conditions (e.g., \texttt{IF}) and loops (e.g., \texttt{FOREACH}).
    \item Every step generates outputs of data types such as \texttt{integer} or \texttt{boolean}, which can be used in subsequent steps.
    \item Steps have inputs that can refer to database tables, columns, and values. For example, the \textit{Look Up Incident Records} step in Figure \ref{fig:sample_flow} matches rows in the \texttt{incident} table where the column \texttt{assigned\_to} has the same value as the \texttt{user} record from the trigger step.
\end{itemize}

The simplest approach to generate low-code workflows is to use prompt engineering on an off-the-shelf LLM. This would be satisfactory if the steps were high-level descriptions and the logic elements were simple. But to obtain reasonable results in this task, we had to decompose it into sub-tasks \cite{wies2023subtask} and use Retrieval-Augmented Generation (RAG) \cite{gao2023retrieval} at various points in the pipeline. We found out that, while prompting an LLM gives reasonable results, we can do better by fine-tuning an SLM.

Our approach was to create a small yet representative training dataset for the sub-tasks comprising \textit{Flow Generation} and fine-tune Mistral-Nemo-12B-Base \cite{mistral2024nemo}. For evaluation, we labeled about a thousand workflows coming from ten domains (around 100 per domain). We also created a special dataset by inviting expert enterprise users to interact with the tool and collecting data generated from their usage. We compare our approach against an instruction fine-tuned SLM as the baseline, as well as against a variety of closed and open-source LLMs, showing the benefits of fine-tuning. As our task is domain-specific, we devised a metric called \textit{Flow Similarity} (\texttt{FlowSim}), a version of tree edit distance that represents workflows as trees.

Our contributions are the following:

\begin{itemize}
    \item We present a case study for how we built an application that generates low-code workflows based on textual user requirements.
    \item We demonstrate that fine-tuning an SLM gives better results than prompting much larger LLMs in this domain-specific task.
    \item We introduce a process to conduct systematic error analysis that reveals model limitations and complements our metrics.
\end{itemize}

\section{Related Work}
\label{sec:related-work}

Prompting \cite{mccann2018natural, brown2020language} has emerged as the predominant method for using LLMs, significantly reducing the effort required compared to fine-tuning \cite{sanh2021multitask, wei2021finetuned}. However, for domain-specific applications requiring extensive prior knowledge, practitioners often encounter limitations such as constrained context windows \cite{peng2023yarn}, which hinder the model’s ability to capture the full scope of the task. To address this, alternative approaches such as RAG \cite{lewis2020retrieval} or domain-specific fine-tuning \cite{gururangan2020don} are employed to incorporate relevant knowledge. In various industry-specific settings, specialized models have demonstrated superior performance over state-of-the-art general-domain LLMs \cite{wu2023bloomberggpt, tu2024towards, colombo2024saullm}. We add to this body of work by showing that a fine-tuned SLM performs better than prompting LLMs in low-code workflow generation.

The domain of \textit{workflows} has been extensively explored in previous work (\citealp{zeng2023flowmind, ayala-bechard-2024-reducing, wornow2024automating, li2024autoflow}, \textit{inter alia}). Notably, \citet{bassamzadeh2024comparative} compare an approach using RAG against fine-tuning a model for workflow generation, while \citet{fan2024workflowllm} propose a synthetic data generation pipeline to enhance generalization. In our case, we combine RAG with fine-tuning an SLM to obtain the highest possible quality.

Evaluation methodologies vary across tasks, with each task having its own quantitative metric for model quality but also its own qualitative error analysis approach \cite{dou2024s, vilar-etal-2006-error, gauthier-melancon-etal-2022-azimuth, bolya2020tide}. In this work, we outline an error analysis approach for a structured output task and leverage it to gain deeper insights into model failure modes.

\section{Methodology}
\label{sec:methodology}
Because workflows are complex structured outputs requiring diverse data in their steps and inputs, we designed a pipeline that relies on RAG and generates the workflow iteratively by asking the LLM to solve sub-tasks.

\subsection{Task Decomposition}
In our domain, workflows are represented as JSON documents, which allows us to break the generation of complete workflows in two stages:
\begin{enumerate}
    \item Given a natural language requirement, generate the plan or \textbf{\textit{outline}} of the workflow, identifying the step names, the order of execution and, crucially, extracting an annotation (description) from the requirement for each step.
    \item For every step in the outline, use the annotation to gather the necessary data from the environment and generate the \textbf{\textit{step inputs}}.
\end{enumerate}

The glue between the two stages are annotations, which also serve as an explanation for the generated step. They allow the model to fill step details, as long as these details are provided by the user. When we retrieve data such as table or column names, these annotations are a key part of the search input.

\begin{figure}[htb!]
  \centering
  \includegraphics[width=\linewidth]{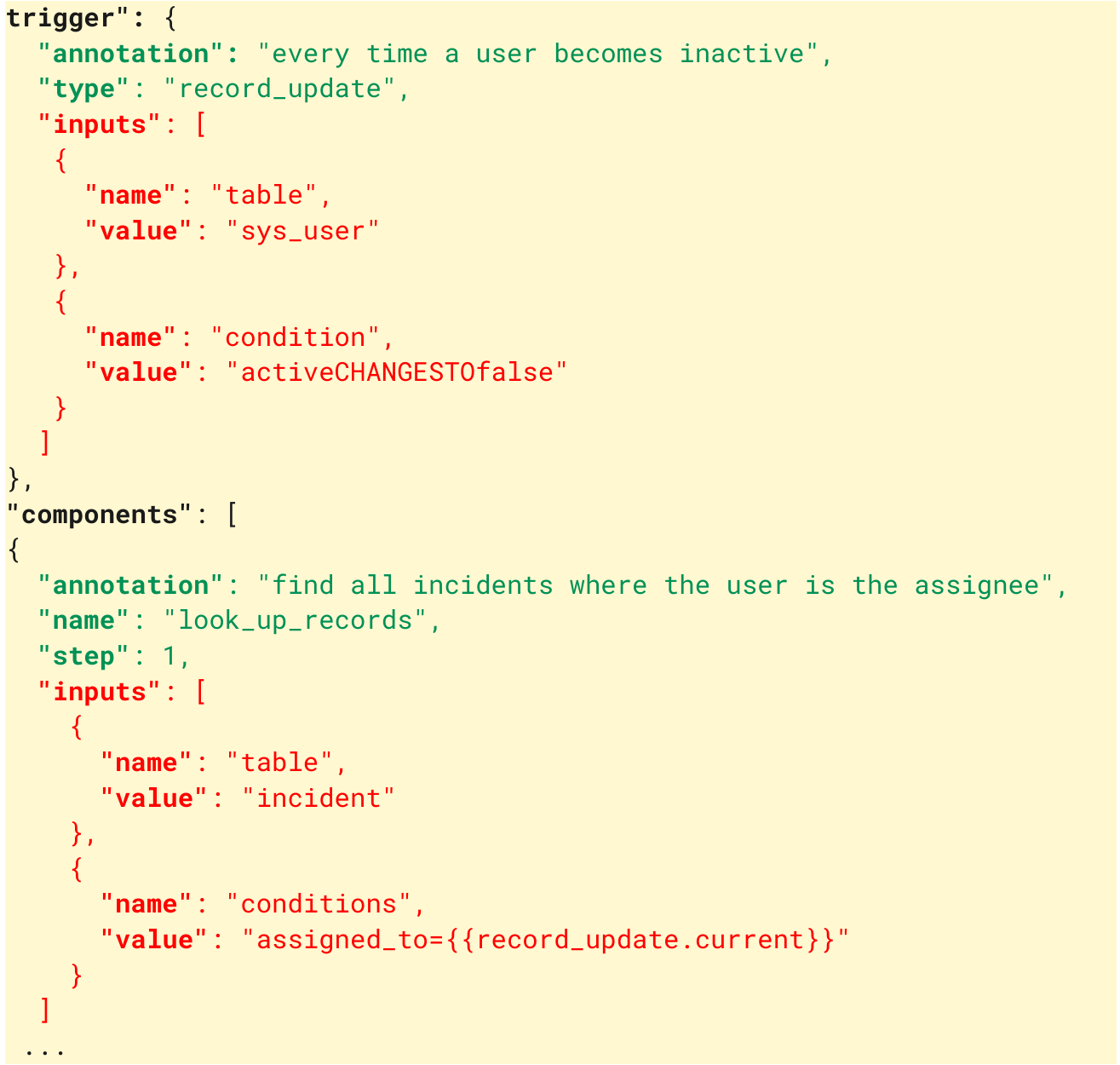}
  \vspace{-0.4cm}
\caption{Sample JSON representation of the trigger and first component steps for the workflow in Figure \ref{fig:sample_flow}.}
\Description{Sample JSON representation of the trigger and first component steps for a workflow}
\label{fig:sample_json}
\vspace{-0.5cm}
\end{figure}

Figure \ref{fig:sample_json} shows the JSON representation of the trigger and first component steps for the workflow in Figure \ref{fig:sample_flow}. The lines in green are generated in the first stage, as part of the outline, and the step inputs in red are generated as part of the second stage. 

\subsection{Retrieval-Augmented Generation}
It has been shown that RAG is necessary to reduce hallucinations when generating structured outputs \cite{ayala-bechard-2024-reducing, bassamzadeh2024comparative}. In our pipeline, RAG is used to suggest the following to the LLM:

\begin{itemize}
    \item Steps to use in the outline. Since the available steps vary across installations of the system, this includes custom steps added by users.
    \item Table names, column names, and values used in steps that read and modify database records in the system. These names and values are used in step inputs.
\end{itemize}

For example, referring to Figure \ref{fig:sample_json}, step names such as \texttt{record\allowbreak\_update} and \texttt{look\_up\_records} can be retrieved based on the text \textit{Every time a user becomes inactive, find all incidents where the user is the assignee. Assign them to their manager.}. Then the table name \texttt{sys\_user}, column name \texttt{active}, and column value \texttt{false} are retrieved from \textit{every time a user becomes inactive}.

\subsection{Pipeline}

\begin{figure}[t]
  \centering
    \includegraphics[width=\linewidth]{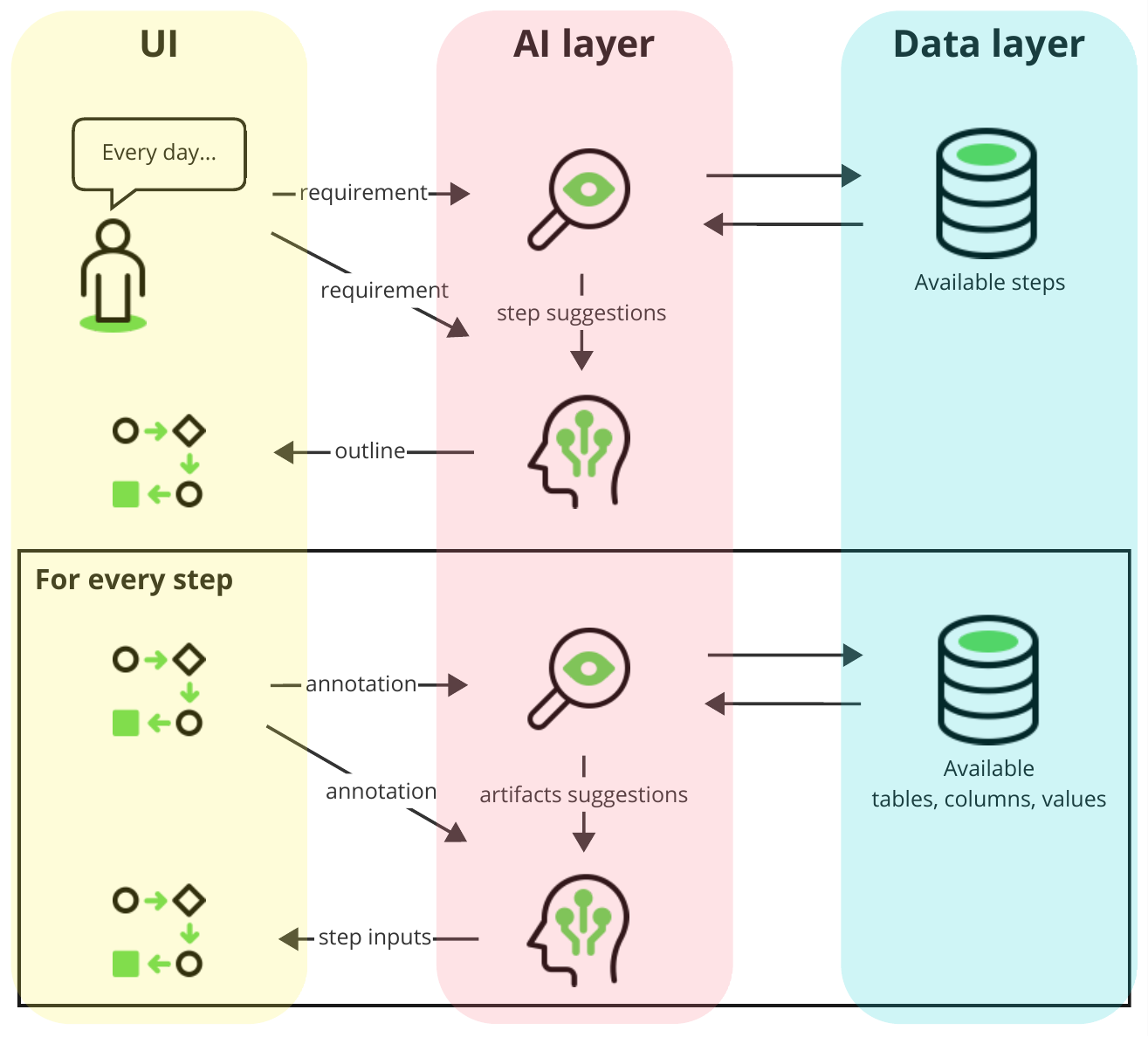}
    \vspace{-0.8cm}
    \caption{System architecture with UI, AI, and data layers.}
    \Description{System architecture with UI, AI, and data layers}
    \label{fig:architecture_diagram}
    \vspace{-0.3cm}
\end{figure}

Figure \ref{fig:architecture_diagram} shows the architecture for the overall pipeline. The UI receives the user requirement and displays the workflow outline and step inputs as they are generated. The AI layer contains the LLM and the retriever. The data layer stores the indexed sources of data, from where the retriever suggests steps and artifacts (tables, columns, values) to the LLM. This layer can be replaced in every installation of the system to allow the LLM to generate output specific to each customer. Note that for a workflow containing $N$ steps, there will be a maximum of $N+1$ LLM calls: one for the outline and one to generate the inputs of each step that accept inputs.

Fine-tuning the SLM and the retriever was done using standard approaches: supervised fine-tuning and contrastive loss training.

\section{Experiments}
\label{sec:experiments}

\subsection{Datasets}
The training dataset comprises two tasks, each corresponding to the two pipeline stages. The first task is \texttt{createFlow}, where the model generates the outline given the natural language requirement and basic workflow metadata. The second task is \texttt{populateInputs}, where given a flow at step $N$, the annotation of step $N$, and any required environment data, the model generates the inputs of this step. In the former, the model receives suggestions for steps while in the latter, it receives suggestions for artifacts that cannot be found deterministically, such as table names that users refer to in the requirement.

We extracted thousands of workflows from internal deployments of the enterprise system for labeling. After discarding those too short or too long, we selected 1,512 workflows. A team was trained to express in natural language a description or requirement of what these workflows accomplish. A downside of using existing workflows is that they are complete while users will create their workflows in draft form and will add complexity iteratively. We therefore created 766 simpler workflows synthetically using domain expertise aiming to cover the data distribution as much as possible. 

For evaluation, we similarly extracted and labeled workflows from 10 customer deployments, around 100 from each. Since each deployment comes from different domains such as retail and banking, the steps and table names differ from those found in the training dataset. We call this dataset the out-of-domain (OOD) set. However, the same downside applies to this evaluation dataset: these are complete workflows. To address this, we invited expert users of the manual system to simulate interacting with the application and generate workflows. Their submitted requirements formed a \textsc{TEST} set of 108 workflows. Table~\ref{tab:dataset_stats} summarizes the number of samples used for training and evaluation.

\begin{table}[hbt!]
  \vspace{-0.3cm}
  \centering
  \caption{Number of examples for \colorbox{training_row!50}{training} (Internal and Synthetic) and \colorbox{eval_row!50}{evaluation} (\textsc{TEST} and OOD).}
  \vspace{-0.2cm}
  \label{tab:dataset_stats}
  \begin{tabular}{lcc}
    \toprule
    \textbf{Dataset} & \textbf{\# \texttt{createFlow}} & \textbf{\# \texttt{populateInputs}} \\
    \midrule
    \rowcolor{training_row!50}
    Internal & 1,512 & 4,709 \\
    \rowcolor{training_row!50}
    Synthetic & 766 & 2,310 \\
    % \midrule
    \rowcolor{eval_row!50}
    \textsc{TEST} & 108 & 367 \\
    \rowcolor{eval_row!50}
    OOD & 1,072 & 4,958 \\
    \bottomrule
  \end{tabular}
  \vspace{-0.5cm}
\end{table}

\begin{figure*}
    \centering
    \includegraphics[width=\linewidth]{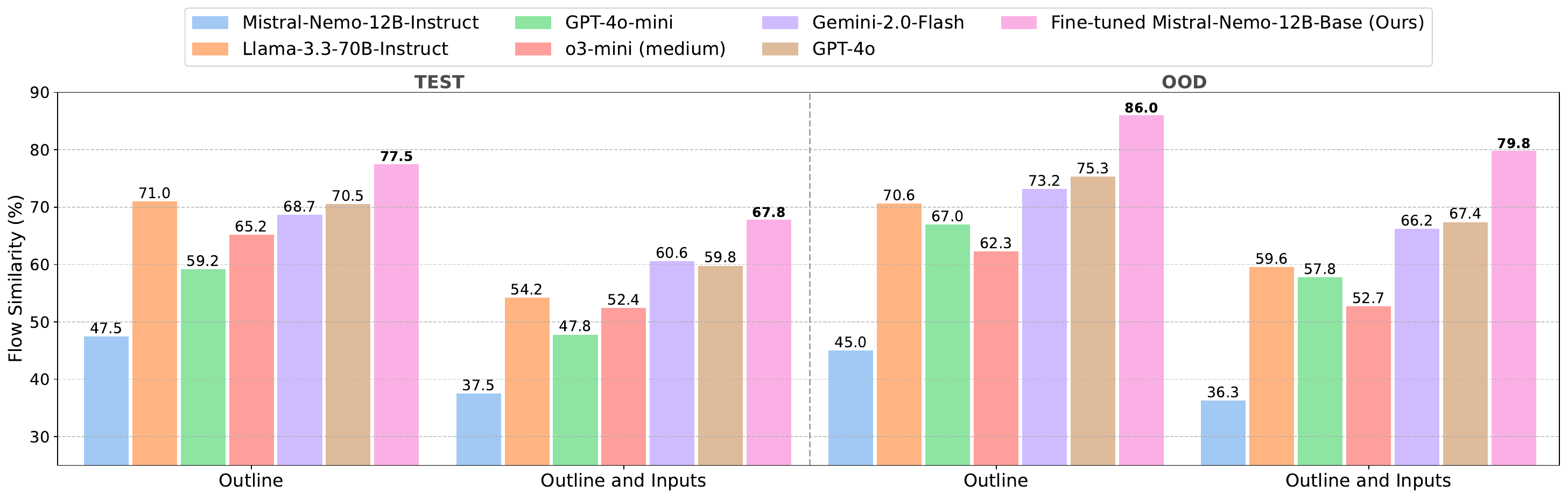}
    \vspace{-0.7cm}
    \caption{Flow similarity results obtained on the \textsc{TEST} and OOD sets for Outline and Outline with inputs.}
    \Description{Flow similarity results obtained on the TEST and OOD sets for Outline and Outline with inputs.}
    \label{fig:flow_sim_results}
    \vspace{-0.3cm}
\end{figure*}

\subsection{Models}
For SLMs, we considered several models such as StarCoderBase-7B \cite{li2023starcoder} and Llama-3.1-8B \cite{grattafiori2024llama}, but we chose Mistral-Nemo-12B-Base \cite{mistral2024nemo} because this model architecture is well-supported and optimized in our enterprise system. However, we obtained similar results with other SLMs as the key ingredient for our fine-tuning is the training data.

As the baseline, we used a model of the same architecture and size that has been instruction fine-tuned but not on data from our workflow domain: Mistral-Nemo-12B-Instruct \cite{mistral2024nemo}. This allows us to validate that domain-specific data is helpful. As for LLMs, we try to cover the current landscape by comparing against:
\begin{itemize}
    \item \textbf{Closed-source:} GPT-4o-mini, GPT-4o, and Gemini-2.0-Flash.
    \item \textbf{Open-source:} Llama-3.3-70B-Instruct \cite{grattafiori2024llama}
    \item \textbf{Reasoning:} o3-mini \cite{openai2025o3mini} with medium reasoning   
\end{itemize}

We crafted two prompt templates, one for each task, to use on all models that were not fine-tuned. Both contain sections for Context, Task definition, Inputs, Guidelines, Constraints with examples, and Output format. Before prompting the LLMs, these templates get populated with data from the system and the retrieved suggestions. In the case of the \texttt{populateInputs} prompt, we dynamically fill the template with instructions according to the inputs to populate (e.g., input types). Table \ref{tab:prompt_stats} gives an idea of the complexity of our prompts by listing the number of input tokens using the \mbox{GPT-4o} tokenizer on the \textsc{TEST} set.

\begin{table}[hbt!]
  \vspace{-0.15cm}
  \caption{Number of tokens in templates and average / standard deviation for prompts sent to the LLMs.}
  \label{tab:prompt_stats}
  \centering
  \vspace{-0.3cm}
  \begin{tabular}{lcc}
    \toprule
    & \textbf{\texttt{createFlow}} & \textbf{\texttt{populateInputs}} \\
    \midrule
    \# tokens in template     & 3,948 & 2,225 \\
    Avg \# prompt tokens      & 5,958 & 3,732 \\
    Std Dev \# prompt tokens  & 282   & 728 \\
    \bottomrule
  \end{tabular}
  \vspace{-0.5cm}
\end{table}

\subsection{Metrics}
We devised our own metric called \textit{Flow Similarity} (\texttt{FlowSim}) to compare generated and expected workflows. As workflows can be represented as trees, similar to how computer programs can be represented as abstract syntax trees, we use the tree edit distance algorithm \cite{doi:10.1137/0218082} to compute a similarity score, where higher is better. We found that this metric correlates well with human evaluation (details in Appendix \ref{sec:appendix_metric}). A labeled expected workflow is required per user requirement, but evaluation can be automated. Appendix \ref{sec:appendix_sample_tree} includes a sample workflow tree.

Our metric, however, has a few limitations. First, we compare against only one workflow whereas there may be several possible versions for the same requirement. Second, we compute exact matches for input strings. Third, it ignores that generated workflows may break basic structure rules (e.g., having an \texttt{ELSE} step without an \texttt{IF}). The first two drawbacks can be addressed by having more than one reference workflow and by using sentence similarity for string inputs, but we leave this to future work. We address the third drawback by computing the percentage of generated examples with at least one such structure error. While the most complete metric is \textbf{\texttt{FlowSim} of outline and inputs}, we also report \textbf{\texttt{FlowSim} of outline}, and the \textbf{percentage of examples with structure errors}.

\section{Results}

Figure \ref{fig:flow_sim_results} shows the results for our fine-tuned SLM and all other LLMs on the small \textsc{TEST} and the larger OOD datasets. Mistral-Nemo-12B-Instruct performs poorly across the board confirming that SLMs require workflow in-domain training data. LLMs of the "mini" sort, even o3-mini with medium reasoning, also do not perform well.

When we prompt larger LLMs, we see competitive results, especially with Gemini-2.0-Flash and \mbox{GPT-4o}. As generating the outline is a simpler task, we see a larger gap between our fine-tuned SLM and these models when we generate the complete workflow (outline and inputs). This gap is 7.2\% on \textsc{TEST} but 12.4\% on OOD, around 10\% if we average the two. All models perform worse when the requirements denote unfinished workflows (\textsc{TEST} set), suggesting a discrepancy between requirements written by expert users and our annotators.

\begin{figure}[htb]
\vspace{-0.2cm}
    \centering
    \includegraphics[width=\linewidth]{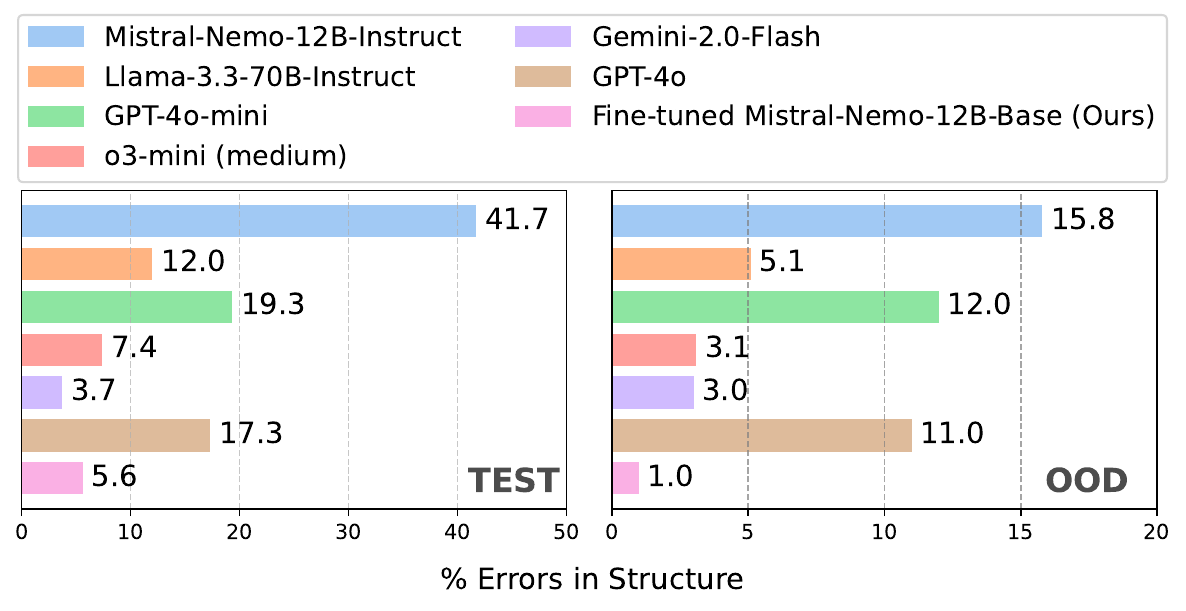}
    \vspace{-0.8cm}
    \caption{Percentage of examples with structure errors on the \textsc{TEST} and OOD sets. Lower is better.}
    \Description{Percentage of examples with structure errors on the TEST and OOD sets. Lower is better.}
    \label{fig:structure_error_results}
    \vspace{-0.3cm}
\end{figure}

While \texttt{FlowSim} values may be high, it could be that the generated workflow had obvious structure errors, which could severely impact usability or require complex post-processing rules. As shown in Figure \ref{fig:structure_error_results}, while \mbox{GPT-4o} performs reasonably well overall, it makes many such errors. With the same prompt, Gemini-2.0-Flash yields even fewer structure errors than our fine-tuned SLM on the \textsc{TEST} set. Refer to Appendix \ref{sec:appendix_results_after_structure_errors} for further analysis.

Our last quantitative comparison helps us determine whether sub-optimal RAG is primarily responsible for errors. Here, we consider only outline generation, as only one retriever call is made to obtain step suggestions, and we consider only the \textsc{TEST} dataset. Table \ref{tab:perfect_rag_results} shows that all models perform better with simulated perfect RAG (i.e., all steps in the expected workflow are part of the suggestions) as opposed to the standard case where the retriever may miss some required steps. Nevertheless, we see that the improvement is only a maximum of 4\%, suggesting that most errors are model errors.

\begin{table}[hbt!]
  \caption{Outline results with perfect RAG (\textsc{TEST} set). Best results are in \textbf{bold} and second best are \underline{underlined}.}
  \label{tab:perfect_rag_results}
  \centering
  \small
  \vspace{-0.2cm}
  \begin{tabular}{lcc}
    \toprule
    \textbf{Model} & \textbf{RAG} & \textbf{Perfect RAG} \\
    \midrule
    Mistral-Nemo-12B-Base (Ours)   & \textbf{77.5}   & \textbf{78.8} \\
    GPT-4o                         & 70.5            & \underline{74.7} \\
    LLama-3.3-70B-Instruct         & \underline{70.6} & 74.4 \\
    Gemini-2.0-Flash               & 68.7            & 70.2 \\
    \bottomrule
  \end{tabular}
  \vspace{-0.4cm}
\end{table}

\section{Error Analysis}
\label{sec:error-analysis}

\subsection{Approach}

To perform a more fine-grained evaluation, our approach to error analysis complements the \texttt{FlowSim} metric by adding interpretability. It makes the process less time-consuming due to its reusability and extensibility.

%Our approach to error analysis preserves the benefits of a qualitative evaluation, complementing and adding interpretability to the \textit{Flow Similarity} metric. It also addresses the drawbacks of this time-consuming, human-driven approach by implementing a reusable and extensible system. 

We began with a qualitative error analysis of our current production model and based on the patterns found, identified features in the ground truth dataset that appeared to negatively impact model output. These findings were then organized into a binary matrix, where 1 indicates a feature's presence in each ground truth sample and 0, an absence (see Table~\ref{tab:binary-matrix}). We then aligned model scores with the matrix, making it easy to identify the features that could be contributing to underperformance.

\begin{table}[hbt!]
\centering
\vspace{-0.2cm}
\caption{Sample binary matrix that characterizes each ground truth sample along \textit{n} feature dimensions, with model scores for each sample.}
\vspace{-0.2cm}
    \begin{tabular}{c|cccc|c}
    \textbf{sample id} & \textbf{feat1} & \textbf{feat2} & ... & \textbf{feat $n$} & \textbf{model score} \\
    \hline
    0 & 0 & 1 & ... & 0 & \cellcolor{Red!70}29 \\
    1 & 1 & 1 & ... & 1 & \cellcolor{Green!30}67 \\
    2 & 0 & 0 & ... & 0 & \cellcolor{Red!50}34 \\
    \vdots & \vdots & \vdots & ... & \vdots & \vdots \\
    $m$ & 0 & 0 & ... & 1 & \cellcolor{Green!70}100
    \end{tabular}
\label{tab:binary-matrix}
\end{table}

This approach greatly expedites error analysis as model comparison is streamlined. New runs and models can be quickly assessed on particular features without reviewing individual samples. If new model runs have poor performance that does not pattern with current features, we can quickly identify items to review for error analysis and potentially add features to the matrix. 

%hybrid approach, starts from manual methods, reading and interpreting errors but gradually becomes more systematic
%facilitates comparison and can be used to understand where a model may have regressed
%generative model, thus have to understand not just what it's doing wrong but what it's doing instead

\subsection{Findings}

We focused our error analysis on the \textsc{TEST} dataset, as it best simulates customer usage of our enterprise application. We identified 24 features across the 108 samples that appeared to impact output quality (see Appendix~\ref{sec:appendix_TEST_features}). We then subdivided the dataset based on sets of related features to better understand each model's ability to (1) create the outline structure, (2) populate inputs, and (3) handle services specific to our enterprise system:

\begin{itemize}
    \item \textsc{\textbf{Structure}} contains structural logic features: \texttt{FOREACH}, \texttt{PARALLEL}, and \texttt{TRY/CATCH}.
    \item \textsc{\textbf{Input}} contains input-related features: \textit{worknotes/descriptions}, \textit{trigger conditions}, and \textit{multiple conditions}.
    \item \textsc{\textbf{Enterprise}} contains features unique to our enterprise system: \textit{service-level agreement (SLA)}, \textit{service catalog}, and \textit{Glide datetime}.
\end{itemize}

We compare our fine-tuned SLM and the LLMs that performed best: Gemini-2.0-Flash and \mbox{GPT-4o}. As Table~\ref{tab:subset-results} shows, our fine-tuned SLM underperforms the LLMs on flows containing structural logic steps. Closer analysis suggests the SLM frequently misses the dependency steps associated with them (e.g. \texttt{FOREACH} is frequently paired with a prior \texttt{look\_up\_records} step, \texttt{PARALLEL} should always consist of more than one branch).

\begin{table}[thb!]
  \vspace{-0.1cm}
  \caption{Model results on three subsets of features, \colorbox{structure_row!50}{\textsc{\textbf{Structure}}}, \colorbox{input_row!50}{\textsc{\textbf{Input}}}, and \colorbox{enterprise_row!50}{\textsc{\textbf{Enterprise}}}. Best results are in \textbf{bold} and second best are \underline{underlined}.}
  \label{tab:subset-results}
  \centering
  \small
  \vspace{-0.2cm}
  \begin{tabular}{lcccc}
    \toprule
    \textbf{Feature} & \textbf{\# Samples} & \textbf{Ours} & \textbf{Gemini} & \textbf{GPT-4o} \\
    \midrule
    \rowcolor{structure_row!50}{\scriptsize \textsc{\textbf{\textcolor{mediumgray}{Structure}}}} & & & & \\
    \rowcolor{structure_row!50}\texttt{FOREACH}   & 22 & 64.1          & \textbf{72.7}   & \underline{71.7} \\
    \rowcolor{structure_row!50}\texttt{PARALLEL}  & 10 & \underline{56.0} & \textbf{65.7}   & 55.1 \\
    \rowcolor{structure_row!50}\texttt{TRY/CATCH} & 5  & \underline{62.2} & \textbf{68.6}   & 50.4 \\
    \rowcolor{structure_row!50}\textit{Average}   &    & 63.8           & \textbf{70.0}   & \underline{64.5} \\
    \midrule
    \rowcolor{input_row!50}{\scriptsize \textsc{\textbf{\textcolor{mediumgray}{Input}}}} & & & & \\
    \rowcolor{input_row!50}Worknotes  & 26 & \textbf{66.8} & 63    & \underline{64.6} \\
    \rowcolor{input_row!50}Triggers   & 37 & \textbf{70.0} & 60.4  & \underline{63.9} \\
    \rowcolor{input_row!50}Multiple   & 23 & \textbf{60.1} & \underline{55.0} & 54.3 \\
    \rowcolor{input_row!50}\textit{Average} & & \textbf{67.3} & 59.8 & \underline{59.9} \\
    \midrule
    \rowcolor{enterprise_row!50}{\scriptsize \textsc{\textbf{\textcolor{mediumgray}{Enterprise}}}} & & & & \\
    \rowcolor{enterprise_row!50}Service Catalog & 7  & \textbf{63.6} & \underline{58.9} & 51.0 \\
    \rowcolor{enterprise_row!50}SLA             & 5  & \textbf{69.6} & \underline{58.8} & 52.2 \\      
    \rowcolor{enterprise_row!50}Glide           & 5  & \textbf{68.6} & \underline{67.8} & 62.2 \\
    \rowcolor{enterprise_row!50}\textit{Average} &   & \textbf{66.8} & \underline{61.6} & 54.6 \\
    \bottomrule
  \end{tabular}
  \vspace{-0.4cm}
\end{table}

The fine-tuned SLM nevertheless consistently outperforms the LLMs on the remaining two subdivisions. The largest margin is on the \textsc{\textbf{Enterprise}} set, where the fine-tuned SLM's average \texttt{FlowSim} score is 12.16\% higher than \mbox{GPT-4o}'s. The smallest margin is also on \textsc{\textbf{Enterprise}}, where the fine-tuned SLM surpasses Gemini-2.0-Flash by 5.35\%. We suspect that learning by demonstrations is more efficient than including complex instructions in a prompt due to the intricacies and specifics of the workflow domain.

Lastly, we observed that workflow steps and conditions were often expressed \underline{implicitly} in the \textsc{TEST} dataset. The requirement \textit{lookup incident tasks and close them}, for example, entails a \texttt{FOREACH} and \texttt{update} step that are not stated overtly. Our results indicate that the fine-tuned SLM is far better than the LLMs on such examples, with a \texttt{FlowSim} score of 65.1 to Gemini's 57.6 and \mbox{GPT-4o}'s 58.5. This indicates the value of labeling as these sort of examples were part of the labeling instructions derived from expectations of how the application would be used.

\section{Conclusion}
We present a case study for building an enterprise generative AI application called \textit{Flow Generation}, which translates textual user requirements into low-code workflows. While prompting state-of-the-art LLMs yield reasonable results, quality can be significantly improved by fine-tuning an SLM. To complement our quantitative metrics, we performed systematic error analysis that reveals strengths and weaknesses of the fine-tuned SLM and the best-performing LLMs. Future work includes improving our custom metric and addressing the gaps identified by our error analysis approach.

%incorporate future work into the conclusion. more data for flowlogics (parallel data comprises just 2.85\% of the training set) ...  We also performed a qualitative evaluation using an adaptable approach to error analysis that increased the explainability and interpretability of our results while being sustainable and reusable for future evaluation.This is especially beneficial in industry where the state-of-the-art is always changing. As demonstrated in our own analysis, our approach also allows results to be decomposed into subsets to drive more granular analysis and identify areas for future work.

% \clearpage

%\section*{Acknowledgments}
%No acknowledgments to give us more space

% Bibliography entries for the entire Anthology, followed by custom entries
%\bibliography{anthology,custom}
% Custom bibliography entries only
\bibliographystyle{ACM-Reference-Format}
\bibliography{custom}

%%% -*-BibTeX-*-
%%% Do NOT edit. File created by BibTeX with style
%%% ACM-Reference-Format-Journals [18-Jan-2012].

\begin{thebibliography}{32}

%%% ====================================================================
%%% NOTE TO THE USER: you can override these defaults by providing
%%% customized versions of any of these macros before the \bibliography
%%% command.  Each of them MUST provide its own final punctuation,
%%% except for \shownote{} and \showURL{}.  The latter two
%%% do not use final punctuation, in order to avoid confusing it with
%%% the Web address.
%%%
%%% To suppress output of a particular field, define its macro to expand
%%% to an empty string, or better, \unskip, like this:
%%%
%%% \newcommand{\showURL}[1]{\unskip}   % LaTeX syntax
%%%
%%% \def \showURL #1{\unskip}           % plain TeX syntax
%%%
%%% ====================================================================

\ifx \showCODEN    \undefined \def \showCODEN     #1{\unskip}     \fi
\ifx \showISBNx    \undefined \def \showISBNx     #1{\unskip}     \fi
\ifx \showISBNxiii \undefined \def \showISBNxiii  #1{\unskip}     \fi
\ifx \showISSN     \undefined \def \showISSN      #1{\unskip}     \fi
\ifx \showLCCN     \undefined \def \showLCCN      #1{\unskip}     \fi
\ifx \shownote     \undefined \def \shownote      #1{#1}          \fi
\ifx \showarticletitle \undefined \def \showarticletitle #1{#1}   \fi
\ifx \showURL      \undefined \def \showURL       {\relax}        \fi
% The following commands are used for tagged output and should be
% invisible to TeX
\providecommand\bibfield[2]{#2}
\providecommand\bibinfo[2]{#2}
\providecommand\natexlab[1]{#1}
\providecommand\showeprint[2][]{arXiv:#2}

\bibitem[Anthropic(2024)]%
        {Claude3S}
\bibfield{author}{\bibinfo{person}{Anthropic}.} \bibinfo{year}{2024}\natexlab{}.
\newblock \bibinfo{title}{The Claude 3 Model Family: Opus, Sonnet, Haiku}.
\newblock \bibinfo{howpublished}{\url{https://www.anthropic.com/news/claude-3-opus-sonnet-haiku}}.
\newblock
\newblock
\shownote{Model card}.


\bibitem[Ayala and Bechard(2024)]%
        {ayala-bechard-2024-reducing}
\bibfield{author}{\bibinfo{person}{Orlando Ayala} {and} \bibinfo{person}{Patrice Bechard}.} \bibinfo{year}{2024}\natexlab{}.
\newblock \showarticletitle{Reducing hallucination in structured outputs via Retrieval-Augmented Generation}. In \bibinfo{booktitle}{\emph{Proceedings of the 2024 Conference of the North American Chapter of the Association for Computational Linguistics: Human Language Technologies (Volume 6: Industry Track)}}, \bibfield{editor}{\bibinfo{person}{Yi~Yang}, \bibinfo{person}{Aida Davani}, \bibinfo{person}{Avi Sil}, {and} \bibinfo{person}{Anoop Kumar}} (Eds.). \bibinfo{publisher}{Association for Computational Linguistics}, \bibinfo{address}{Mexico City, Mexico}, \bibinfo{pages}{228--238}.
\newblock
\href{https://doi.org/10.18653/v1/2024.naacl-industry.19}{doi:\nolinkurl{10.18653/v1/2024.naacl-industry.19}}


\bibitem[Bassamzadeh and Methani(2024)]%
        {bassamzadeh2024comparative}
\bibfield{author}{\bibinfo{person}{Nastaran Bassamzadeh} {and} \bibinfo{person}{Chhaya Methani}.} \bibinfo{year}{2024}\natexlab{}.
\newblock \showarticletitle{A Comparative Study of DSL Code Generation: Fine-Tuning vs. Optimized Retrieval Augmentation}.
\newblock \bibinfo{journal}{\emph{arXiv preprint arXiv:2407.02742}} (\bibinfo{year}{2024}).
\newblock


\bibitem[Bolya et~al\mbox{.}(2020)]%
        {bolya2020tide}
\bibfield{author}{\bibinfo{person}{Daniel Bolya}, \bibinfo{person}{Sean Foley}, \bibinfo{person}{James Hays}, {and} \bibinfo{person}{Judy Hoffman}.} \bibinfo{year}{2020}\natexlab{}.
\newblock \showarticletitle{Tide: A general toolbox for identifying object detection errors}. In \bibinfo{booktitle}{\emph{Computer Vision--ECCV 2020: 16th European Conference, Glasgow, UK, August 23--28, 2020, Proceedings, Part III 16}}. Springer, \bibinfo{pages}{558--573}.
\newblock


\bibitem[Brown et~al\mbox{.}(2020)]%
        {brown2020language}
\bibfield{author}{\bibinfo{person}{Tom Brown}, \bibinfo{person}{Benjamin Mann}, \bibinfo{person}{Nick Ryder}, \bibinfo{person}{Melanie Subbiah}, \bibinfo{person}{Jared~D Kaplan}, \bibinfo{person}{Prafulla Dhariwal}, \bibinfo{person}{Arvind Neelakantan}, \bibinfo{person}{Pranav Shyam}, \bibinfo{person}{Girish Sastry}, \bibinfo{person}{Amanda Askell}, {et~al\mbox{.}}} \bibinfo{year}{2020}\natexlab{}.
\newblock \showarticletitle{Language models are few-shot learners}.
\newblock \bibinfo{journal}{\emph{Advances in neural information processing systems}}  \bibinfo{volume}{33} (\bibinfo{year}{2020}), \bibinfo{pages}{1877--1901}.
\newblock


\bibitem[Colombo et~al\mbox{.}(2024)]%
        {colombo2024saullm}
\bibfield{author}{\bibinfo{person}{Pierre Colombo}, \bibinfo{person}{Telmo Pires}, \bibinfo{person}{Malik Boudiaf}, \bibinfo{person}{Rui Melo}, \bibinfo{person}{Dominic Culver}, \bibinfo{person}{Etienne Malaboeuf}, \bibinfo{person}{Gabriel Hautreux}, \bibinfo{person}{Johanne Charpentier}, {and} \bibinfo{person}{Michael Desa}.} \bibinfo{year}{2024}\natexlab{}.
\newblock \showarticletitle{Saullm-54b \& saullm-141b: Scaling up domain adaptation for the legal domain}.
\newblock \bibinfo{journal}{\emph{Advances in Neural Information Processing Systems}}  \bibinfo{volume}{37} (\bibinfo{year}{2024}), \bibinfo{pages}{129672--129695}.
\newblock


\bibitem[Dou et~al\mbox{.}(2024)]%
        {dou2024s}
\bibfield{author}{\bibinfo{person}{Shihan Dou}, \bibinfo{person}{Haoxiang Jia}, \bibinfo{person}{Shenxi Wu}, \bibinfo{person}{Huiyuan Zheng}, \bibinfo{person}{Weikang Zhou}, \bibinfo{person}{Muling Wu}, \bibinfo{person}{Mingxu Chai}, \bibinfo{person}{Jessica Fan}, \bibinfo{person}{Caishuang Huang}, \bibinfo{person}{Yunbo Tao}, {et~al\mbox{.}}} \bibinfo{year}{2024}\natexlab{}.
\newblock \showarticletitle{What's Wrong with Your Code Generated by Large Language Models? An Extensive Study}.
\newblock \bibinfo{journal}{\emph{arXiv preprint arXiv:2407.06153}} (\bibinfo{year}{2024}).
\newblock


\bibitem[Fan et~al\mbox{.}(2024)]%
        {fan2024workflowllm}
\bibfield{author}{\bibinfo{person}{Shengda Fan}, \bibinfo{person}{Xin Cong}, \bibinfo{person}{Yuepeng Fu}, \bibinfo{person}{Zhong Zhang}, \bibinfo{person}{Shuyan Zhang}, \bibinfo{person}{Yuanwei Liu}, \bibinfo{person}{Yesai Wu}, \bibinfo{person}{Yankai Lin}, \bibinfo{person}{Zhiyuan Liu}, {and} \bibinfo{person}{Maosong Sun}.} \bibinfo{year}{2024}\natexlab{}.
\newblock \showarticletitle{WorkflowLLM: Enhancing Workflow Orchestration Capability of Large Language Models}.
\newblock \bibinfo{journal}{\emph{arXiv preprint arXiv:2411.05451}} (\bibinfo{year}{2024}).
\newblock


\bibitem[Fu et~al\mbox{.}(2024)]%
        {fu-etal-2024-tiny}
\bibfield{author}{\bibinfo{person}{Xue-Yong Fu}, \bibinfo{person}{Md~Tahmid~Rahman Laskar}, \bibinfo{person}{Elena Khasanova}, \bibinfo{person}{Cheng Chen}, {and} \bibinfo{person}{Shashi Tn}.} \bibinfo{year}{2024}\natexlab{}.
\newblock \showarticletitle{Tiny Titans: Can Smaller Large Language Models Punch Above Their Weight in the Real World for Meeting Summarization?}. In \bibinfo{booktitle}{\emph{Proceedings of the 2024 Conference of the North American Chapter of the Association for Computational Linguistics: Human Language Technologies (Volume 6: Industry Track)}}, \bibfield{editor}{\bibinfo{person}{Yi~Yang}, \bibinfo{person}{Aida Davani}, \bibinfo{person}{Avi Sil}, {and} \bibinfo{person}{Anoop Kumar}} (Eds.). \bibinfo{publisher}{Association for Computational Linguistics}, \bibinfo{address}{Mexico City, Mexico}, \bibinfo{pages}{387--394}.
\newblock
\href{https://doi.org/10.18653/v1/2024.naacl-industry.33}{doi:\nolinkurl{10.18653/v1/2024.naacl-industry.33}}


\bibitem[Gao et~al\mbox{.}(2023)]%
        {gao2023retrieval}
\bibfield{author}{\bibinfo{person}{Yunfan Gao}, \bibinfo{person}{Yun Xiong}, \bibinfo{person}{Xinyu Gao}, \bibinfo{person}{Kangxiang Jia}, \bibinfo{person}{Jinliu Pan}, \bibinfo{person}{Yuxi Bi}, \bibinfo{person}{Yi Dai}, \bibinfo{person}{Jiawei Sun}, \bibinfo{person}{Qianyu Guo}, \bibinfo{person}{Meng Wang}, {et~al\mbox{.}}} \bibinfo{year}{2023}\natexlab{}.
\newblock \showarticletitle{Retrieval-Augmented Generation for Large Language Models: A Survey}.
\newblock \bibinfo{journal}{\emph{CoRR}} (\bibinfo{year}{2023}).
\newblock


\bibitem[Gauthier-melancon et~al\mbox{.}(2022)]%
        {gauthier-melancon-etal-2022-azimuth}
\bibfield{author}{\bibinfo{person}{Gabrielle Gauthier-melancon}, \bibinfo{person}{Orlando Marquez~Ayala}, \bibinfo{person}{Lindsay Brin}, \bibinfo{person}{Chris Tyler}, \bibinfo{person}{Frederic Branchaud-charron}, \bibinfo{person}{Joseph Marinier}, \bibinfo{person}{Karine Grande}, {and} \bibinfo{person}{Di Le}.} \bibinfo{year}{2022}\natexlab{}.
\newblock \showarticletitle{Azimuth: Systematic Error Analysis for Text Classification}. In \bibinfo{booktitle}{\emph{Proceedings of the 2022 Conference on Empirical Methods in Natural Language Processing: System Demonstrations}}, \bibfield{editor}{\bibinfo{person}{Wanxiang Che} {and} \bibinfo{person}{Ekaterina Shutova}} (Eds.). \bibinfo{publisher}{Association for Computational Linguistics}, \bibinfo{address}{Abu Dhabi, UAE}, \bibinfo{pages}{298--310}.
\newblock
\href{https://doi.org/10.18653/v1/2022.emnlp-demos.30}{doi:\nolinkurl{10.18653/v1/2022.emnlp-demos.30}}


\bibitem[Gemini et~al\mbox{.}(2023)]%
        {team2023gemini}
\bibfield{author}{\bibinfo{person}{Gemini}, \bibinfo{person}{Rohan Anil}, \bibinfo{person}{Sebastian Borgeaud}, \bibinfo{person}{Jean-Baptiste Alayrac}, \bibinfo{person}{Jiahui Yu}, \bibinfo{person}{Radu Soricut}, \bibinfo{person}{Johan Schalkwyk}, \bibinfo{person}{Andrew~M Dai}, \bibinfo{person}{Anja Hauth}, \bibinfo{person}{Katie Millican}, {et~al\mbox{.}}} \bibinfo{year}{2023}\natexlab{}.
\newblock \showarticletitle{Gemini: a family of highly capable multimodal models}.
\newblock \bibinfo{journal}{\emph{arXiv preprint arXiv:2312.11805}} (\bibinfo{year}{2023}).
\newblock


\bibitem[Grattafiori et~al\mbox{.}(2024)]%
        {grattafiori2024llama}
\bibfield{author}{\bibinfo{person}{Aaron Grattafiori}, \bibinfo{person}{Abhimanyu Dubey}, \bibinfo{person}{Abhinav Jauhri}, \bibinfo{person}{Abhinav Pandey}, \bibinfo{person}{Abhishek Kadian}, \bibinfo{person}{Ahmad Al-Dahle}, \bibinfo{person}{Aiesha Letman}, \bibinfo{person}{Akhil Mathur}, \bibinfo{person}{Alan Schelten}, \bibinfo{person}{Alex Vaughan}, {et~al\mbox{.}}} \bibinfo{year}{2024}\natexlab{}.
\newblock \showarticletitle{The llama 3 herd of models}.
\newblock \bibinfo{journal}{\emph{arXiv preprint arXiv:2407.21783}} (\bibinfo{year}{2024}).
\newblock


\bibitem[Gururangan et~al\mbox{.}(2020)]%
        {gururangan2020don}
\bibfield{author}{\bibinfo{person}{Suchin Gururangan}, \bibinfo{person}{Ana Marasovi{\'c}}, \bibinfo{person}{Swabha Swayamdipta}, \bibinfo{person}{Kyle Lo}, \bibinfo{person}{Iz Beltagy}, \bibinfo{person}{Doug Downey}, {and} \bibinfo{person}{Noah~A Smith}.} \bibinfo{year}{2020}\natexlab{}.
\newblock \showarticletitle{Don't stop pretraining: Adapt language models to domains and tasks}.
\newblock \bibinfo{journal}{\emph{arXiv preprint arXiv:2004.10964}} (\bibinfo{year}{2020}).
\newblock


\bibitem[Hurst et~al\mbox{.}(2024)]%
        {hurst2024gpt}
\bibfield{author}{\bibinfo{person}{Aaron Hurst}, \bibinfo{person}{Adam Lerer}, \bibinfo{person}{Adam~P Goucher}, \bibinfo{person}{Adam Perelman}, \bibinfo{person}{Aditya Ramesh}, \bibinfo{person}{Aidan Clark}, \bibinfo{person}{AJ Ostrow}, \bibinfo{person}{Akila Welihinda}, \bibinfo{person}{Alan Hayes}, \bibinfo{person}{Alec Radford}, {et~al\mbox{.}}} \bibinfo{year}{2024}\natexlab{}.
\newblock \showarticletitle{Gpt-4o system card}.
\newblock \bibinfo{journal}{\emph{arXiv preprint arXiv:2410.21276}} (\bibinfo{year}{2024}).
\newblock


\bibitem[Lewis et~al\mbox{.}(2020)]%
        {lewis2020retrieval}
\bibfield{author}{\bibinfo{person}{Patrick Lewis}, \bibinfo{person}{Ethan Perez}, \bibinfo{person}{Aleksandra Piktus}, \bibinfo{person}{Fabio Petroni}, \bibinfo{person}{Vladimir Karpukhin}, \bibinfo{person}{Naman Goyal}, \bibinfo{person}{Heinrich K{\"u}ttler}, \bibinfo{person}{Mike Lewis}, \bibinfo{person}{Wen-tau Yih}, \bibinfo{person}{Tim Rockt{\"a}schel}, {et~al\mbox{.}}} \bibinfo{year}{2020}\natexlab{}.
\newblock \showarticletitle{Retrieval-augmented generation for knowledge-intensive nlp tasks}.
\newblock \bibinfo{journal}{\emph{Advances in neural information processing systems}}  \bibinfo{volume}{33} (\bibinfo{year}{2020}), \bibinfo{pages}{9459--9474}.
\newblock


\bibitem[Li et~al\mbox{.}(2023)]%
        {li2023starcoder}
\bibfield{author}{\bibinfo{person}{Raymond Li}, \bibinfo{person}{Loubna~Ben Allal}, \bibinfo{person}{Yangtian Zi}, \bibinfo{person}{Niklas Muennighoff}, \bibinfo{person}{Denis Kocetkov}, \bibinfo{person}{Chenghao Mou}, \bibinfo{person}{Marc Marone}, \bibinfo{person}{Christopher Akiki}, \bibinfo{person}{Jia Li}, \bibinfo{person}{Jenny Chim}, {et~al\mbox{.}}} \bibinfo{year}{2023}\natexlab{}.
\newblock \showarticletitle{Starcoder: may the source be with you!}
\newblock \bibinfo{journal}{\emph{arXiv preprint arXiv:2305.06161}} (\bibinfo{year}{2023}).
\newblock


\bibitem[Li et~al\mbox{.}(2024)]%
        {li2024autoflow}
\bibfield{author}{\bibinfo{person}{Zelong Li}, \bibinfo{person}{Shuyuan Xu}, \bibinfo{person}{Kai Mei}, \bibinfo{person}{Wenyue Hua}, \bibinfo{person}{Balaji Rama}, \bibinfo{person}{Om Raheja}, \bibinfo{person}{Hao Wang}, \bibinfo{person}{He Zhu}, {and} \bibinfo{person}{Yongfeng Zhang}.} \bibinfo{year}{2024}\natexlab{}.
\newblock \showarticletitle{Autoflow: Automated workflow generation for large language model agents}.
\newblock \bibinfo{journal}{\emph{arXiv preprint arXiv:2407.12821}} (\bibinfo{year}{2024}).
\newblock


\bibitem[McCann et~al\mbox{.}(2018)]%
        {mccann2018natural}
\bibfield{author}{\bibinfo{person}{Bryan McCann}, \bibinfo{person}{Nitish~Shirish Keskar}, \bibinfo{person}{Caiming Xiong}, {and} \bibinfo{person}{Richard Socher}.} \bibinfo{year}{2018}\natexlab{}.
\newblock \showarticletitle{The natural language decathlon: Multitask learning as question answering}.
\newblock \bibinfo{journal}{\emph{arXiv preprint arXiv:1806.08730}} (\bibinfo{year}{2018}).
\newblock


\bibitem[Mistral(2024)]%
        {mistral2024nemo}
\bibfield{author}{\bibinfo{person}{Mistral}.} \bibinfo{year}{2024}\natexlab{}.
\newblock \bibinfo{title}{Mistral Nemo}.
\newblock
\urldef\tempurl%
\url{https://mistral.ai/news/mistral-nemo}
\showURL{%
\tempurl}
\newblock
\shownote{Accessed: 2024}.


\bibitem[OpenAI(2025)]%
        {openai2025o3mini}
\bibfield{author}{\bibinfo{person}{OpenAI}.} \bibinfo{year}{2025}\natexlab{}.
\newblock \bibinfo{title}{OpenAI o3-mini System Card}.
\newblock \bibinfo{howpublished}{https://cdn.openai.com/o3-mini-system-card-feb10.pdf}.
\newblock
\newblock
\shownote{Accessed: March 19, 2025}.


\bibitem[Peng et~al\mbox{.}(2023)]%
        {peng2023yarn}
\bibfield{author}{\bibinfo{person}{Bowen Peng}, \bibinfo{person}{Jeffrey Quesnelle}, \bibinfo{person}{Honglu Fan}, {and} \bibinfo{person}{Enrico Shippole}.} \bibinfo{year}{2023}\natexlab{}.
\newblock \showarticletitle{Yarn: Efficient context window extension of large language models}.
\newblock \bibinfo{journal}{\emph{arXiv preprint arXiv:2309.00071}} (\bibinfo{year}{2023}).
\newblock


\bibitem[Sanh et~al\mbox{.}(2021)]%
        {sanh2021multitask}
\bibfield{author}{\bibinfo{person}{Victor Sanh}, \bibinfo{person}{Albert Webson}, \bibinfo{person}{Colin Raffel}, \bibinfo{person}{Stephen~H Bach}, \bibinfo{person}{Lintang Sutawika}, \bibinfo{person}{Zaid Alyafeai}, \bibinfo{person}{Antoine Chaffin}, \bibinfo{person}{Arnaud Stiegler}, \bibinfo{person}{Teven~Le Scao}, \bibinfo{person}{Arun Raja}, {et~al\mbox{.}}} \bibinfo{year}{2021}\natexlab{}.
\newblock \showarticletitle{Multitask prompted training enables zero-shot task generalization}.
\newblock \bibinfo{journal}{\emph{arXiv preprint arXiv:2110.08207}} (\bibinfo{year}{2021}).
\newblock


\bibitem[Trad and Chehab(2024)]%
        {make6010018}
\bibfield{author}{\bibinfo{person}{Fouad Trad} {and} \bibinfo{person}{Ali Chehab}.} \bibinfo{year}{2024}\natexlab{}.
\newblock \showarticletitle{Prompt Engineering or Fine-Tuning? A Case Study on Phishing Detection with Large Language Models}.
\newblock \bibinfo{journal}{\emph{Machine Learning and Knowledge Extraction}} \bibinfo{volume}{6}, \bibinfo{number}{1} (\bibinfo{year}{2024}), \bibinfo{pages}{367--384}.
\newblock
\showISSN{2504-4990}
\href{https://doi.org/10.3390/make6010018}{doi:\nolinkurl{10.3390/make6010018}}


\bibitem[Tu et~al\mbox{.}(2024)]%
        {tu2024towards}
\bibfield{author}{\bibinfo{person}{Tao Tu}, \bibinfo{person}{Shekoofeh Azizi}, \bibinfo{person}{Danny Driess}, \bibinfo{person}{Mike Schaekermann}, \bibinfo{person}{Mohamed Amin}, \bibinfo{person}{Pi-Chuan Chang}, \bibinfo{person}{Andrew Carroll}, \bibinfo{person}{Charles Lau}, \bibinfo{person}{Ryutaro Tanno}, \bibinfo{person}{Ira Ktena}, {et~al\mbox{.}}} \bibinfo{year}{2024}\natexlab{}.
\newblock \showarticletitle{Towards generalist biomedical AI}.
\newblock \bibinfo{journal}{\emph{Nejm Ai}} \bibinfo{volume}{1}, \bibinfo{number}{3} (\bibinfo{year}{2024}), \bibinfo{pages}{AIoa2300138}.
\newblock


\bibitem[Vilar et~al\mbox{.}(2006)]%
        {vilar-etal-2006-error}
\bibfield{author}{\bibinfo{person}{David Vilar}, \bibinfo{person}{Jia Xu}, \bibinfo{person}{Luis~Fernando D{'}Haro}, {and} \bibinfo{person}{Hermann Ney}.} \bibinfo{year}{2006}\natexlab{}.
\newblock \showarticletitle{Error Analysis of Statistical Machine Translation Output}. In \bibinfo{booktitle}{\emph{Proceedings of the Fifth International Conference on Language Resources and Evaluation ({LREC}`06)}}, \bibfield{editor}{\bibinfo{person}{Nicoletta Calzolari}, \bibinfo{person}{Khalid Choukri}, \bibinfo{person}{Aldo Gangemi}, \bibinfo{person}{Bente Maegaard}, \bibinfo{person}{Joseph Mariani}, \bibinfo{person}{Jan Odijk}, {and} \bibinfo{person}{Daniel Tapias}} (Eds.). \bibinfo{publisher}{European Language Resources Association (ELRA)}, \bibinfo{address}{Genoa, Italy}.
\newblock
\urldef\tempurl%
\url{https://aclanthology.org/L06-1244/}
\showURL{%
\tempurl}


\bibitem[Wei et~al\mbox{.}(2021)]%
        {wei2021finetuned}
\bibfield{author}{\bibinfo{person}{Jason Wei}, \bibinfo{person}{Maarten Bosma}, \bibinfo{person}{Vincent~Y Zhao}, \bibinfo{person}{Kelvin Guu}, \bibinfo{person}{Adams~Wei Yu}, \bibinfo{person}{Brian Lester}, \bibinfo{person}{Nan Du}, \bibinfo{person}{Andrew~M Dai}, {and} \bibinfo{person}{Quoc~V Le}.} \bibinfo{year}{2021}\natexlab{}.
\newblock \showarticletitle{Finetuned language models are zero-shot learners}.
\newblock \bibinfo{journal}{\emph{arXiv preprint arXiv:2109.01652}} (\bibinfo{year}{2021}).
\newblock


\bibitem[Wies et~al\mbox{.}(2023)]%
        {wies2023subtask}
\bibfield{author}{\bibinfo{person}{Noam Wies}, \bibinfo{person}{Yoav Levine}, {and} \bibinfo{person}{Amnon Shashua}.} \bibinfo{year}{2023}\natexlab{}.
\newblock \showarticletitle{Sub-Task Decomposition Enables Learning in Sequence to Sequence Tasks}. In \bibinfo{booktitle}{\emph{The Eleventh International Conference on Learning Representations}}.
\newblock
\urldef\tempurl%
\url{https://openreview.net/forum?id=BrJATVZDWEH}
\showURL{%
\tempurl}


\bibitem[Wornow et~al\mbox{.}(2024)]%
        {wornow2024automating}
\bibfield{author}{\bibinfo{person}{Michael Wornow}, \bibinfo{person}{Avanika Narayan}, \bibinfo{person}{Krista Opsahl-Ong}, \bibinfo{person}{Quinn McIntyre}, \bibinfo{person}{Nigam Shah}, {and} \bibinfo{person}{Christopher Re}.} \bibinfo{year}{2024}\natexlab{}.
\newblock \showarticletitle{Automating the enterprise with foundation models}.
\newblock \bibinfo{journal}{\emph{Proceedings of the VLDB Endowment}} \bibinfo{volume}{17}, \bibinfo{number}{11} (\bibinfo{year}{2024}), \bibinfo{pages}{2805--2812}.
\newblock


\bibitem[Wu et~al\mbox{.}(2023)]%
        {wu2023bloomberggpt}
\bibfield{author}{\bibinfo{person}{Shijie Wu}, \bibinfo{person}{Ozan Irsoy}, \bibinfo{person}{Steven Lu}, \bibinfo{person}{Vadim Dabravolski}, \bibinfo{person}{Mark Dredze}, \bibinfo{person}{Sebastian Gehrmann}, \bibinfo{person}{Prabhanjan Kambadur}, \bibinfo{person}{David Rosenberg}, {and} \bibinfo{person}{Gideon Mann}.} \bibinfo{year}{2023}\natexlab{}.
\newblock \showarticletitle{Bloomberggpt: A large language model for finance}.
\newblock \bibinfo{journal}{\emph{arXiv preprint arXiv:2303.17564}} (\bibinfo{year}{2023}).
\newblock


\bibitem[Zeng et~al\mbox{.}(2023)]%
        {zeng2023flowmind}
\bibfield{author}{\bibinfo{person}{Zhen Zeng}, \bibinfo{person}{William Watson}, \bibinfo{person}{Nicole Cho}, \bibinfo{person}{Saba Rahimi}, \bibinfo{person}{Shayleen Reynolds}, \bibinfo{person}{Tucker Balch}, {and} \bibinfo{person}{Manuela Veloso}.} \bibinfo{year}{2023}\natexlab{}.
\newblock \showarticletitle{FlowMind: automatic workflow generation with LLMs}. In \bibinfo{booktitle}{\emph{Proceedings of the Fourth ACM International Conference on AI in Finance}}. \bibinfo{pages}{73--81}.
\newblock


\bibitem[Zhang and Shasha(1989)]%
        {doi:10.1137/0218082}
\bibfield{author}{\bibinfo{person}{Kaizhong Zhang} {and} \bibinfo{person}{Dennis Shasha}.} \bibinfo{year}{1989}\natexlab{}.
\newblock \showarticletitle{Simple Fast Algorithms for the Editing Distance between Trees and Related Problems}.
\newblock \bibinfo{journal}{\emph{SIAM J. Comput.}} \bibinfo{volume}{18}, \bibinfo{number}{6} (\bibinfo{year}{1989}), \bibinfo{pages}{1245--1262}.
\newblock
\href{https://doi.org/10.1137/0218082}{doi:\nolinkurl{10.1137/0218082}}
\showeprint{https://doi.org/10.1137/0218082}


\end{thebibliography}

% \clearpage
\appendix

\section{Correlation of Flow Similarity with Human Evaluation}
\label{sec:appendix_metric}

To validate that our custom metric indicated that a generated workflow satisfied the user requirement, we compared it with human scores. We selected 30 random samples from the \textsc{TEST} dataset, generated workflows using our fine-tuned SLM, and asked domain experts in our Quality Engineering department to assign scores from 0 to 10 to each result. A score of zero meant that the generated workflow does not at all represent the user requirement and a score of ten meant a perfect generated workflow.

\begin{table}[hbt!]
    \centering
    \begin{tabular}{lccc}
        \hline 
        & \textbf{Outline} & \textbf{Outline and inputs} \\
        \hline
        Pearson &  0.78 {\footnotesize (4.4e-7)}  & 0.78 {\footnotesize (2.9e-7)}   \\
        Spearman &  0.83 {\footnotesize (1.5e-8)}  & 0.76 {\footnotesize (9.1e-7)}  \\
        \hline
    \end{tabular}
    \caption{Correlation between human evaluation scores and \texttt{FlowSim} metrics. P-value is shown in parentheses.}
    \label{tab:correlation_metric}
\end{table}

We performed correlation tests for the outline only as well as for outline and inputs, computing both Pearson and Spearman correlation coefficients along with their associated p-values. As Table \ref{tab:correlation_metric} shows, the correlation coefficients are higher than 75\% in all cases and the p-values are very small, confirming that the correlation numbers are statistically significant. This gives us confidence that the \texttt{FlowSim} metric is appropriate for our use case.

\section{Example of representing a workflow as a tree}
\label{sec:appendix_sample_tree}

\begin{figure}[htbp]
  \centering
    \includegraphics[width=\linewidth]{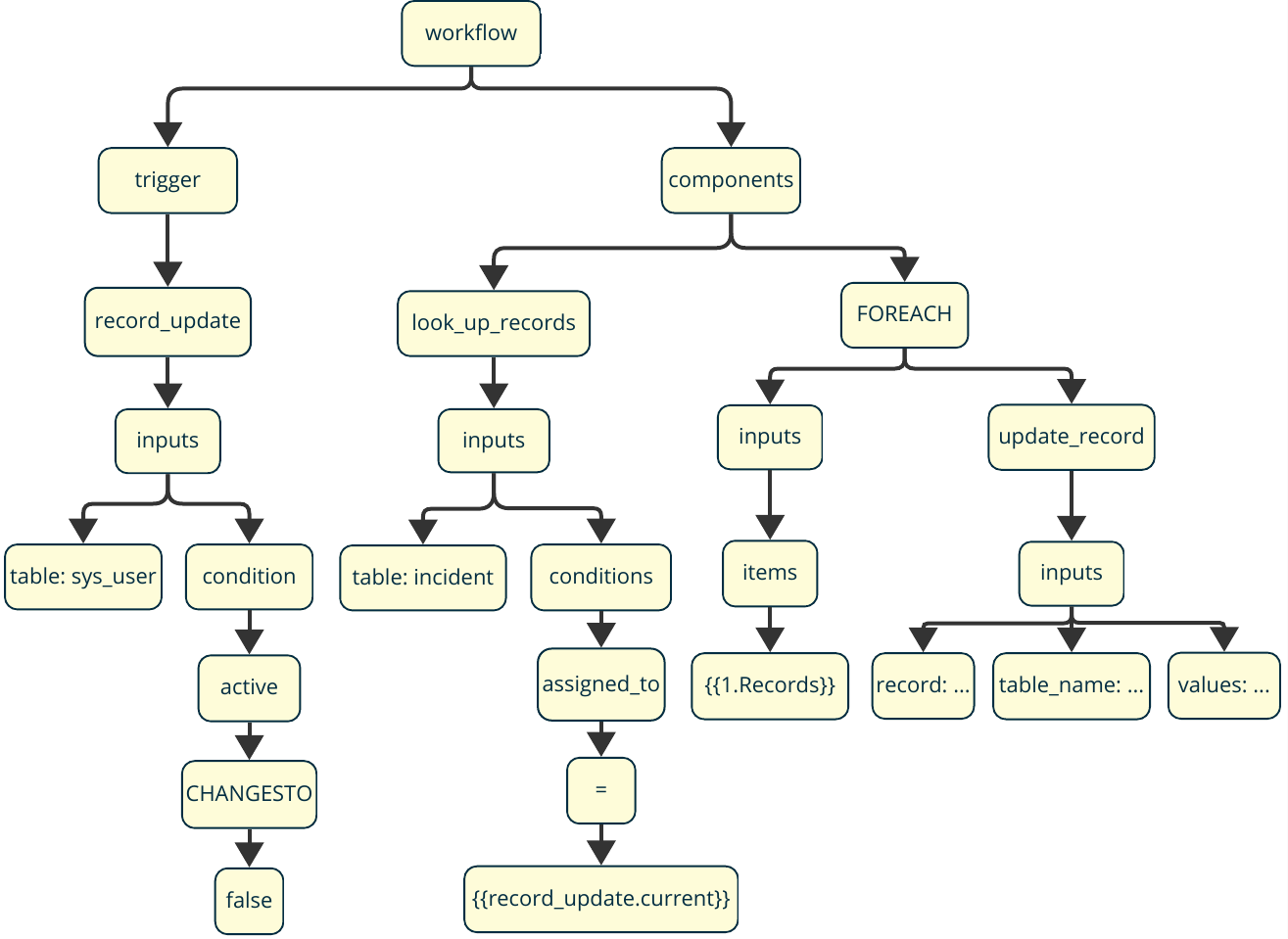}
    \caption{Workflow in Figure \ref{fig:sample_flow} represented as a tree.}
    \Description{Workflow represented as a tree.}
    \label{fig:flow_as_tree}
\end{figure}

The tree in Figure \ref{fig:flow_as_tree} is the representation of the workflow that can be generated from the requirement \textit{Every time a user becomes inactive, find all
incidents where the user is the assignee.
Assign them to their manager}. This workflow is also shown in Figure \ref{fig:sample_flow}.

\section{Flow Similarity Results with Structure Validation}
\label{sec:appendix_results_after_structure_errors}

As Figure \ref{fig:structure_error_results} shows, many LLMs generate workflows that have obvious structure errors. For instance, on the \textsc{TEST} set, 17.3\% of the 108 workflows generated by \mbox{GPT-4o}, and 12\% of those generated by Llama-3.3-70B-Instruct have such errors.

While some of these errors can be corrected via post-processing, it requires additional engineering effort. The best model would be the one that minimizes this percentage. To take into account this failure mode, we computed another set of metrics where all these "broken" workflows received an Outline score of zero, as these are unusable. Consequently, the score for Outline and inputs also becomes zero.

\begin{figure*}[t]
    \centering
    \includegraphics[width=\linewidth]{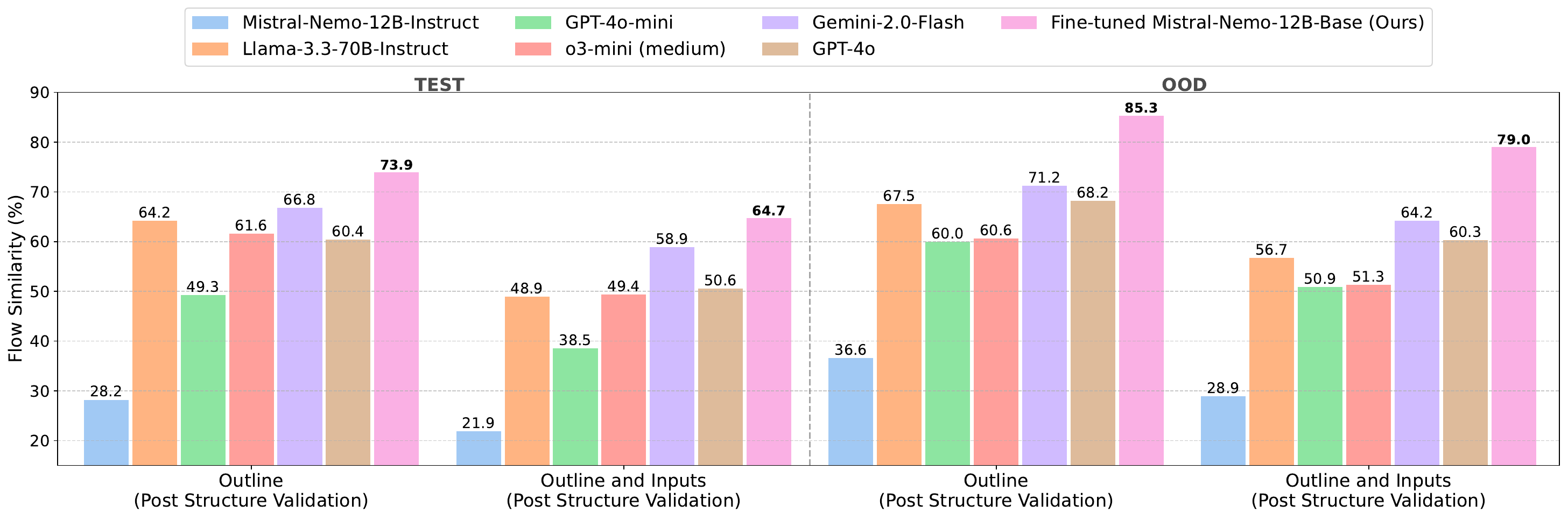}
    \caption{\texttt{FlowSim} results obtained on the \textsc{TEST} and OOD sets for Outline and Outline with inputs, after structure validation is applied.}
    \Description{FlowSim results obtained on the TEST and OOD sets for Outline and Outline with inputs, after structure validation is applied.}
    \label{fig:flow_sim_results_struct}
\end{figure*}

Figure \ref{fig:flow_sim_results_struct} shows the \texttt{FlowSim} results obtained for all models after this structure validation is applied. While our fine-tuned SLM still gives significant improvements over prompting LLMs, we observe that \mbox{GPT-4o} suffers a  substantial degradation compared to Gemini-2.0-Flash. Excluding structure errors, both of these LLMs performed virtually the same.

%\section{Human Annotators}

%\textcolor{red}{TODO: add section about human annotators and the overall labeling process. who is labeling? (contractors vs FTE), how is the labeling and reviewing process being done? how much time does it take to label examples? any demographic info about our labelers? what was the training process?}

\section{\textsc{TEST} Dataset Features}
\label{sec:appendix_TEST_features}
A total of 24 features were identified during the qualitative error analysis. They are presented in Table~\ref{tab:single_column_features}, grouped by theme. Some elements were not included in any training data or were rare in the training data. These, we grouped to address training gaps in the future. Others, including flow logics, are common in training data and have regular patterning of model behavior. They are useful groupings to check model performance at a granular level without reviewing individual samples. Finally, some of the \textsc{TEST} samples are adversarial and are particularly useful for assessing default behavior and seeing if anything `breaks' the model.

\begin{table*}[t]
  \centering
  \small
  \caption{Features identified during the error analysis process, grouped by type.}
  \label{tab:single_column_features}
  \begin{tabular}{l}
    \toprule
    \textbf{Features by Type} \\
    \midrule
    \textsc{\textbf{\footnotesize \textcolor{mediumgray}{Not included in training}}} \\
    \textbf{Triggers}: null, service catalog \\
    \textbf{Actions}: get catalog variables, ask for approval \\
    \textbf{Inputs}: glide query, dynamic ME \\
    \addlinespace[0.5em]
    \textsc{\textbf{\footnotesize \textcolor{mediumgray}{Uncommon in training data}}} \\
    \textbf{Triggers}: SLA, conditionals, weekly \\
    \textbf{Inputs}: requestor-related \\
    \addlinespace[0.5em]
    \textsc{\textbf{\footnotesize \textcolor{mediumgray}{Flow Logics}}} \\
    \texttt{DOUNTIL}, \texttt{PARALLEL}, \texttt{FOREACH}, \texttt{TRY-CATCH} \\
    \addlinespace[0.5em]
    \textsc{\textbf{\footnotesize \textcolor{mediumgray}{Common}}} \\
    \textbf{Actions}: MS Teams / Slack message, notifications / emails, log, worknotes / descriptions \\
    \textbf{Inputs}: Complex inputs \\
    \addlinespace[0.5em]
    \textsc{\textbf{\footnotesize \textcolor{mediumgray}{Difficult / Edge Cases}}} \\
    Out of scope, ambiguous, implicit actions, misleading input, incorrect / incomplete \\
    \bottomrule
  \end{tabular}
\end{table*}

\end{document}